\documentclass{article}

\usepackage{times}
\usepackage{graphicx} 

\usepackage{natbib}

\usepackage{algorithm}
\usepackage{algorithmic}
\usepackage{amsthm}

\usepackage{hyperref}

\usepackage{url}
\usepackage{caption}
\usepackage{subcaption}
\usepackage{amsmath}
\usepackage[accepted]{icml2017}

\newcommand{\stitle}[1]{\vspace{5pt} \noindent\textbf{#1.}\ }

\newtheorem{theorem}{Theorem}

\newtheorem{proposition}{Proposition}
\newtheorem{remark}{Remark}
\newtheorem{definition}{Definition}

\newcommand{\esm}[1]{\ensuremath{#1}}

\newcommand{\ms}[1]{\esm{\mathsf{#1}}}

\newcommand\reals{\ms{R}}

\newcommand\sparam{\alpha}

\newcommand\pathfn{\gamma}

\newcommand\synteq{::=}

\newcommand\integratedgrads{\ms{IntegratedGrads}}
\newcommand\pathintegratedgrads{\ms{PathIntegratedGrads}}
\newcommand\relu{\ms{ReLU}}
\newcommand\sigmoid{\ms{Sigmoid}}
\newcommand\xbase{x'}

\begin{document}

\twocolumn[

\icmltitle{Axiomatic Attribution for Deep Networks}
\icmlsetsymbol{equal}{*}

\begin{icmlauthorlist}
\icmlauthor{Mukund Sundararajan}{equal,google}
\icmlauthor{Ankur Taly}{equal,google}
\icmlauthor{Qiqi Yan}{equal,google}
\end{icmlauthorlist}

\icmlaffiliation{google}{Google Inc., Mountain View, USA}

\icmlcorrespondingauthor{Mukund Sundararajan}{mukunds@google.com}
\icmlcorrespondingauthor{Ankur Taly}{ataly@google.com}

\icmlkeywords{Deep Learning, Explanations, Gradients, Attribution}

\vskip 0.3in
]




\printAffiliationsAndNotice{\icmlEqualContribution} 

\begin{abstract}
  We study the problem of attributing the prediction of a deep network
  to its input features, a problem previously studied by several other
  works. We identify two fundamental axioms---\emph{Sensitivity} and
  \emph{Implementation Invariance} that attribution methods ought to
  satisfy. We show that they are not satisfied by most known
  attribution methods, which we consider to be a fundamental weakness
  of those methods. We use the axioms to guide the design of a new
  attribution method called \emph{Integrated Gradients}. Our method
  requires no modification to the original network and is extremely
  simple to implement; it just needs a few calls to the standard
  gradient operator. We apply this method to a couple of image models,
  a couple of text models and a chemistry model, demonstrating its
  ability to debug networks, to extract rules from a network, and
  to enable users to engage with models better.
\end{abstract}

\section{Motivation and Summary of Results}\label{sec:intro}

We study the problem of \emph{attributing the prediction of a deep
  network to its input features}.
\begin{definition}
  Formally, suppose we have a function $F: \reals^n \rightarrow [0,1]$
  that represents a deep network, and an input $x = (x_1,\ldots,x_n) \in \reals^n$.
  An attribution of the prediction at input $x$ relative to a baseline
  input $\xbase$ is a vector $A_F(x, \xbase) = (a_1,\ldots,a_n) \in \reals^n$
  where $a_i$ is the \emph{contribution} of $x_i$ to the
  prediction $F(x)$.
\end{definition}
For instance, in an object
recognition network, an attribution method could tell us which pixels
of the image were responsible for a certain label being picked (see
Figure~\ref{fig:intgrad-finalgrad}). The attribution problem was
previously studied by various papers
~\cite{BSHKHM10,SVZ13,SGSK16,BMBMS16,SDBR14}.

The intention of these works is to understand the input-output
behavior of the deep network, which gives us the ability to improve
it.  Such understandability is critical to all computer programs,
including machine learning models. There are also other applications
of attribution. They could be used within a product driven by machine
learning to provide a rationale for the recommendation. For instance,
a deep network that predicts a condition based on imaging could help
inform the doctor of the part of the image that resulted in the
recommendation. This could help the doctor understand the strengths
and weaknesses of a model and compensate for it. We give such an
example in Section~\ref{sec:app-dr}. Attributions could also be used
by developers in an exploratory sense. For instance, we could use a
deep network to extract insights that could be then used in a rule-based
system. In Section~\ref{sec:app-qc}, we give such an example.

A significant challenge in designing an attribution technique is that
they are hard to evaluate empirically. As we discuss in
Section~\ref{sec:uniqueness}, it is hard to tease apart errors that
stem from the misbehavior of the model versus the misbehavior of the
attribution method. To compensate for this shortcoming, we take an
axiomatic approach. In Section~\ref{sec:two-axioms} we identify two
axioms that every attribution method must satisfy. Unfortunately
most previous methods do not satisfy one of these two axioms.
In Section~\ref{sec:method}, we use the axioms to identify a new method,
called \textbf{integrated gradients}.

Unlike previously proposed methods, integrated gradients do not need
any instrumentation of the network, and can be computed easily using
a few calls to the gradient operation, allowing even novice
practitioners to easily apply the technique.

In Section~\ref{sec:app}, we demonstrate the ease of applicability over
several deep networks, including two images networks, two
text processing networks, and a chemistry network. These
applications demonstrate the use of our technique in either improving
our understanding of the network, performing debugging, performing rule
extraction, or aiding an end user in understanding the network's
prediction.

\begin{remark}
  \label{rem:baseline}
  Let us briefly examine the need for the baseline in the definition of
  the attribution problem.
  A common way for humans to perform attribution relies on counterfactual intuition.
  When we assign blame to a certain cause we implicitly consider the absence of
  the cause as a baseline for comparing outcomes. In a deep network, we model
  the absence using a single baseline input. For most deep networks, a natural
  baseline exists in the input space where the prediction is neutral. For instance,
  in object recognition networks, it is the black image. The need for a baseline
  has also been pointed out by prior work on attribution~\cite{SGSK16, BMBMS16}.
\end{remark}


\section{Two Fundamental Axioms}
\label{sec:two-axioms}

We now discuss two axioms (desirable characteristics) for attribution
methods. We find that other feature attribution methods in literature
break at least one of the two axioms.  These methods include DeepLift~\cite{SGSK16, SGK17},
Layer-wise relevance propagation (LRP)
\cite{BMBMS16}, Deconvolutional networks~\cite{ZF14}, and Guided
back-propagation~\cite{SDBR14}. As we will see in
Section~\ref{sec:method}, these axioms will also guide the design of
our method.

\stitle{Gradients}
For linear models, ML practitioners regularly inspect the products of the
model coefficients and the feature values in order to debug predictions.
Gradients (of the output with respect
to the input) is a natural analog of the model coefficients for a deep network,
and therefore the product of the gradient and feature values  is a reasonable
starting point for an attribution method ~\cite{BSHKHM10, SVZ13}; see the third column of
Figure~\ref{fig:intgrad-finalgrad} for examples. The problem with
gradients is that they break \emph{sensitivity}, a property
that all attribution methods should satisfy.

\subsection{Axiom: Sensitivity(a)} An attribution method satisfies \emph{Sensitivity(a)}
if for every input and baseline that differ in one feature but have different
predictions then the differing feature should be given a non-zero attribution.
(Later in the paper, we will have a part (b) to this definition.)

Gradients violate Sensitivity(a): For a concrete example, consider a
one variable, one ReLU network, $f(x) = 1 - \relu(1-x)$.  Suppose the
baseline is $x=0$ and the input is $x=2$. The function changes from $0$ to
$1$, but because $f$ becomes flat at $x=1$, the gradient method gives
attribution of $0$ to $x$.  Intuitively, gradients break Sensitivity
because the prediction function may flatten at the input and thus have
zero gradient despite the function value at the input being different
from that at the baseline. This phenomenon has been reported in
previous work~\cite{SGSK16}.

Practically, the lack of sensitivity causes gradients to focus on
irrelevant features (see the ``fireboat'' example in
Figure~\ref{fig:intgrad-finalgrad}).

\stitle{Other back-propagation based approaches} A second set of approaches
involve back-propagating the final prediction score through each layer
of the network down to the individual features.  These include
DeepLift, Layer-wise relevance propagation (LRP), Deconvolutional networks (DeConvNets),
and Guided back-propagation. These methods differ in
the specific backpropagation logic for various activation functions
(e.g., ReLU, MaxPool, etc.).

Unfortunately, Deconvolution networks (DeConvNets), and
Guided back-propagation violate Sensitivity(a).
This is because these methods back-propogate through a ReLU node
only if the ReLU is turned on at the input. This makes the method
similar to gradients, in that, the attribution is zero for features
with zero gradient at the input despite a non-zero gradient at the
baseline. We defer the specific counterexamples to Appendix~\ref{sec:examples}.

Methods like DeepLift and LRP tackle the Sensitivity issue by
employing a baseline, and in some sense try to compute ``discrete
gradients'' instead of (instantaeneous) gradients at the input.
(The two methods differ in the specifics of how they compute the discrete gradient).
But the idea is that a large, discrete step will avoid flat regions, avoiding a
breakage of sensitivity.  Unfortunately, these methods violate a
different requirement on attribution methods.

\subsection{Axiom: Implementation Invariance} Two networks are \emph{functionally
equivalent} if their outputs are equal for all inputs, despite
having very different implementations. Attribution
methods should satisfy \emph{Implementation Invariance}, i.e., the
attributions are always identical for two functionally equivalent
networks.  To motivate this, notice that attribution can be
colloquially defined as assigning the blame (or credit) for the
output to the input features. Such a definition does not refer to
implementation details.

We now discuss intuition for why DeepLift and LRP break Implementation
Invariance; a concrete example is provided in Appendix~\ref{sec:examples}.

First, notice that gradients are invariant to implementation.
In fact, the chain-rule for
gradients $\frac{\partial f}{\partial g}=\frac{\partial f}{\partial
  h}\cdot \frac{\partial h}{\partial g}$ is essentially about
implementation invariance. To see this, think of $g$ and $f$ as the input
and output of a system, and $h$ being some implementation detail of the
system.  The gradient of output $f$ to input $g$
can be computed either directly by $\frac{\partial f}{\partial g}$,
ignoring the intermediate function $h$ (implementation detail), or by
invoking the chain rule via $h$. This is exactly how backpropagation works.

Methods like LRP and DeepLift replace gradients with discrete
gradients and still use a modified form of backpropagation to compose
discrete gradients into attributions.
Unfortunately, the chain rule does not hold for discrete gradients in
general. Formally
$\frac{f(x_1)-f(x_0)}{g(x_1)-g(x_0)}\neq\frac{f(x_1)-f(x_0)}{h(x_1)-h(x_0)}\cdot\frac{h(x_1)-h(x_0)}{g(x_1)-g(x_0)}$
, and therefore these methods fail to satisfy implementation invariance.

If an attribution method fails to satisfy Implementation Invariance,
the attributions are potentially sensitive to unimportant aspects of
the models.
For instance, if the network architecture has more degrees of freedom
than needed to represent a function then there may be two sets of values
for the network parameters that lead to the same function.
The training procedure can converge at either set of values depending
on the initializtion or for other reasons, but the underlying network
function would remain the same. It is undesirable that attributions differ
for such reasons.

\section{Our Method: Integrated Gradients}
\label{sec:method}

We are now ready to describe our technique. Intuitively, our technique
combines the Implementation Invariance of Gradients along with the Sensitivity
of techniques like LRP or DeepLift.

Formally, suppose we have a
function $F: \reals^n \rightarrow [0,1]$ that represents a deep network.
Specifically, let $x \in \reals^n$ be the input at hand, and
$\xbase \in \reals^n$ be the baseline input. For image networks, the baseline
could be the black image, while for text models it could be the zero
embedding vector.

We consider the straightline path (in $\reals^n$) from the baseline $\xbase$ to the input
$x$, and compute the gradients at all points along the path. Integrated gradients
are obtained by cumulating these gradients. Specifically, integrated gradients
are defined as the path intergral of the gradients along the straightline
path from the baseline $\xbase$ to the input $x$.

The integrated gradient along the $i^{th}$ dimension for an input $x$ and baseline
$\xbase$ is defined as follows. Here,
$\tfrac{\partial F(x)}{\partial x_i}$ is the gradient of
$F(x)$ along the $i^{th}$ dimension.
\begin{equation}
\small
\integratedgrads_i(x) \synteq (x_i-\xbase_i)\times\int_{\sparam=0}^{1} \tfrac{\partial F(\xbase + \sparam\times(x-\xbase))}{\partial x_i  }~d\sparam
\end{equation}
\stitle{Axiom: Completeness}
Integrated gradients satisfy an axiom called \emph{completeness}
that the attributions add
up to the difference between the output of $F$ at the input
$x$ and the \emph{baseline}
$\xbase$. This axiom is identified as being desirable by Deeplift and LRP.
It is a sanity check that the attribution method is somewhat comprehensive
in its accounting, a property that is clearly desirable if the network’s score is
used in a numeric sense, and not just to pick the top label, for e.g.,
a model estimating insurance premiums from credit features of individuals.

This is formalized by the proposition below, which
instantiates the fundamental theorem of calculus for path
integrals.
\begin{proposition}\label{prop:additivity}
  If $F: \reals^n \rightarrow \reals$ is differentiable almost
  everywhere
  \footnote{Formally, this means the function $F$ is continuous everywhere
    and the partial derivative of $F$ along each input dimension satisfies
    Lebesgue's integrability condition, i.e., the set of discontinuous points
    has measure zero. Deep networks built out of Sigmoids, ReLUs, and pooling
    operators satisfy this condition.} then
  $$\Sigma_{i=1}^{n} \integratedgrads_i(x) = F(x) -
  F(\xbase)$$
\end{proposition}

For most deep networks, it is possible to choose a baseline such that
the prediction at the baseline is near zero ($F(\xbase) \approx
0$). (For image models, the black image baseline indeed satisfies
this property.) In such cases, there is an intepretation of the resulting
attributions that ignores the baseline and amounts to distributing the
output to the individual input features.

\begin{remark}\label{rem:compsens}
  Integrated gradients satisfies Sensivity(a) because Completeness implies
Sensivity(a) and is thus a strengthening of the Sensitivity(a) axiom. This
is because Sensitivity(a) refers to a case where the baseline and the input
differ only in one variable, for which Completeness asserts that the difference
in the two output values is equal to the attribution to this variable.
Attributions generated by integrated gradients satisfy Implementation Invariance
since they are based only on the gradients of the function represented
by the network.
\end{remark}

\section{Uniqueness of Integrated Gradients}\label{sec:uniqueness}

Prior literature has relied on empirically evaluating the attribution
technique.  For instance, in the context of an object recognition
task, ~\cite{SBMBM15} suggests that we select the top $k$ pixels by
attribution and randomly vary their intensities and then measure the
drop in score. If the attribution method is good, then the drop in
score should be large. However, the images resulting from pixel
perturbation could be unnatural, and it could be that the scores drop
simply because the network has never seen anything like it in
training.  (This is less of a concern with linear or logistic
models where the simplicity of the model ensures that ablating a
feature does not cause strange interactions.)

A different evaluation technique considers images with human-drawn
bounding boxes around objects, and computes the percentage of pixel
attribution inside the box. While for most objects, one would
expect the pixels located on the object to be most important for the
prediction, in some cases the context in which the object occurs may
also contribute to the prediction. The “cabbage butterfly” image from
Figure~\ref{fig:intgrad-finalgrad} is a good example of this where the
pixels on the leaf are also surfaced by the integrated gradients.

Roughly, we found that every empirical evaluation technique we could
think of could not differentiate between artifacts that stem from
perturbing the data, a misbehaving model, and a misbehaving
attribution method.  This was why we turned to an axiomatic approach
in designing a good attribution method (Section~\ref{sec:two-axioms}).
While our method satisfies Sensitivity and Implementation Invariance,
it certainly isn't the unique method to do so.

We now justify the selection of the integrated gradients method in two
steps.  First, we identify a class of methods called Path methods that
generalize integrated gradients.  We discuss that path methods are the
only methods to satisfy certain desirable axioms.  Second, we argue
why integrated gradients is somehow canonical among the different path
methods.

\subsection{Path Methods}
\label{sec:path_methods}

\begin{figure}
  \centering
\includegraphics[width=0.5\columnwidth]{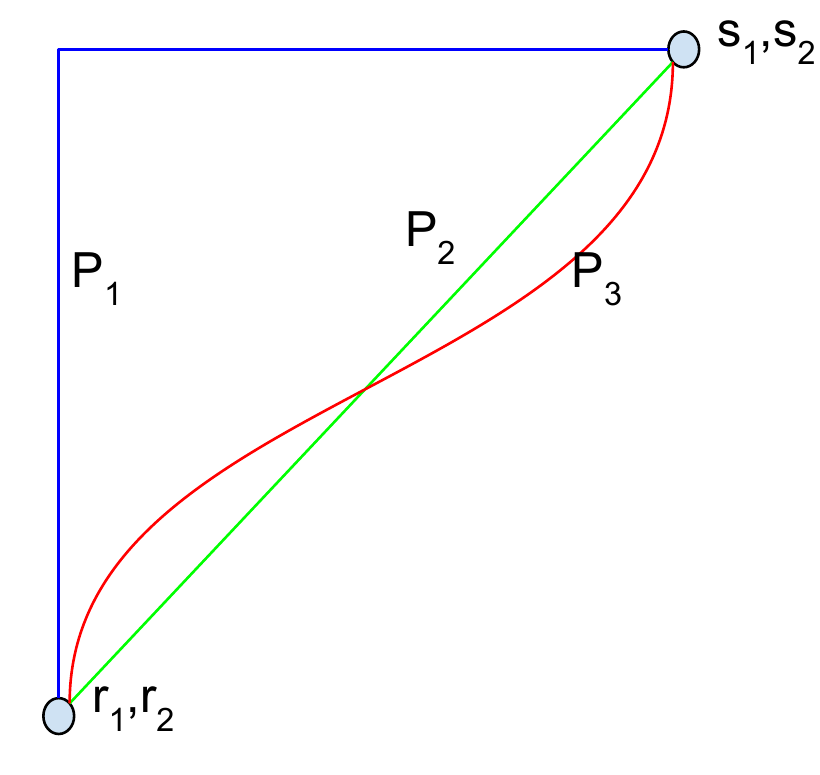}
  \caption{Three paths between an a baseline $(r_1, r_2)$ and an input $(s_1, s_2)$.
    Each path corresponds to a different attribution method. The path $P_2$ corresponds to the path used by integrated gradients.}
  \label{fig:three-paths}
\end{figure}

Integrated gradients aggregate the gradients along the
inputs that fall on the straightline between the baseline and
the input. There are many other
(non-straightline) paths that monotonically interpolate between the
two points, and each such path will yield a different attribution method.  For instance,
consider the simple case when the input is two dimensional.
Figure~\ref{fig:three-paths} has examples of three paths, each of which corresponds to a different
attribution method.

Formally, let $\pathfn = (\pathfn_1, \ldots, \pathfn_n): [0,1]
\rightarrow \reals^n$ be a smooth function specifying a path in
$\reals^n$ from the baseline $\xbase$ to the input $x$, i.e.,
$\pathfn(0) = \xbase$ and $\pathfn(1) = x$.

Given a path function $\pathfn$, \emph{path integrated gradients} are obtained by
integrating the gradients along the path $\pathfn(\sparam)$
for $\sparam \in [0,1]$. Formally, path
integrated gradients along the $i^{th}$ dimension for an input $x$
is defined as follows.
\begin{equation}
\pathintegratedgrads^{\pathfn}_i(x) \synteq \int_{\sparam=0}^{1} \tfrac{\partial F(\pathfn(\sparam))}{\partial \pathfn_i(\sparam)  }~\tfrac{\partial \pathfn_i(\sparam)}{\partial \sparam}  ~d\sparam
\end{equation}
where $\tfrac{\partial F(x)}{\partial x_i}$ is the gradient of
$F$ along the $i^{th}$ dimension at $x$.

Attribution methods based on path integrated gradients are collectively
known as \emph{path methods}. 
Notice that integrated gradients is a path method
for the straightline path specified $\pathfn(\sparam) = \xbase + \sparam\times(x-\xbase)$
for $\sparam \in [0,1]$.

\begin{remark}
  All path methods satisfy Implementation Invariance. This follows from
  the fact that they are defined using the underlying gradients, which
  do not depend on the implementation. They also satisfy Completeness
  (the proof is similar
  to that of Proposition~\ref{prop:additivity})
  and Sensitvity(a) which is implied by Completeness
  (see Remark~\ref{rem:compsens}).
\end{remark}

More interestingly, path methods are the only methods
that satisfy certain desirable axioms. (For formal definitions of the
axioms and proof of Proposition~\ref{prop:path}, see
Friedman~\cite{Friedman}.)

\stitle{Axiom: Sensitivity(b)} (called Dummy in~\cite{Friedman}) If the
function implemented by the deep network does not depend (mathematically)
on some variable, then the attribution to that variable is always
zero.

This is a natural complement to the definition of Sensitivity(a) from
Section~\ref{sec:two-axioms}. This definition captures desired
insensitivity of the attributions.

\stitle{Axiom: Linearity}
Suppose that we linearly composed two deep networks modeled by the
functions $f_1$ and $f_2$ to form a third network that models the
function $a\times f_1 + b\times f_2$, i.e., a linear
combination of the two networks. Then we'd like the attributions for
$a\times f_1 + b\times f_2$ to be the weighted sum of the attributions
for $f_1$ and $f_2$ with weights $a$ and $b$
respectively. Intuitively, we would like the attributions to preserve
any linearity within the network.

  \begin{proposition}(Theorem~1~\cite{Friedman})
    \label{prop:path}
  Path methods are the only
  attribution methods that always satisfy Implementation Invariance,
  Sensitivity(b), Linearity, and Completeness.
\end{proposition}

\begin{remark}
  We note that these path integrated gradients have been used within
  the cost-sharing literature in economics where the function models the
  cost of a project as a function of the
  demands of various participants, and the attributions correspond to
  cost-shares. Integrated gradients correspond to a cost-sharing
  method called Aumann-Shapley~\cite{AS74}.
  Proposition~\ref{prop:path} holds for our attribution problem
  because mathematically the cost-sharing problem corresponds to the
  attribution problem with the benchmark fixed at the zero vector.
  (Implementation Invariance is implicit in the cost-sharing literature
  as the cost functions are considered directly in their mathematical
  form.)
\end{remark}

\subsection{Integrated Gradients is Symmetry-Preserving}

In this section, we formalize why the straightline path chosen by integrated
gradients is canonical.  First, observe that it is the simplest path
that one can define mathematically.  Second, a natural property for
attribution methods is to preserve symmetry, in the following sense.

\stitle{Symmetry-Preserving} Two input variables are symmetric
w.r.t.\ a function if swapping them does not change the function.
For instance, $x$ and $y$ are symmetric w.r.t.\ $F$ if and only if
$F(x, y) = F(y, x)$ for all values of $x$ and $y$.
An attribution method
is symmetry preserving, if for all inputs that have identical values
for symmetric variables and baselines that have identical
values for symmetric variables,  the symmetric variables receive identical
attributions.

E.g., consider the logistic model $\sigmoid(x_1+x_2+\dots)$.
$x_1$ and $x_2$ are symmetric variables for this model. For
an input where $x_1 = x_2 = 1$ (say) and baseline where $x_1 = x_2 = 0$ (say),
a symmetry preserving method must offer identical attributions to $x_1$ and $x_2$.

It seems natural to ask for symmetry-preserving attribution methods because if two
variables play the exact same role in the network (i.e., they are
symmetric and have the same values in the baseline and the input) then they ought to
receive the same attrbiution.

\begin{theorem}\label{thm:unique}
Integrated gradients is the unique path method that is symmetry-preserving.
\end{theorem}
The proof is provided in Appendix~\ref{sec:proof}.
\begin{remark}
If we allow averaging over the attributions from multiple paths, then
are other methods that satisfy all the axioms in
Theorem~\ref{thm:unique}.
In particular, there is the method by Shapley-Shubik~\cite{ShaShu71}
from the cost sharing literature, and used by~\cite{LundbergL16, DSZ16}
to compute feature attributions (though they were not studying deep networks).
In this method, the attribution is the average of those from $n!$ extremal paths; here $n$
is the number of features. Here each such path considers an ordering
of the input features, and sequentially changes the input feature from
its value at the baseline to its value at the input.
This method yields attributions that are different from integrated gradients. If the function of interest is $min(x_1,x_2)$, the baseline is $x_1=x_2=0$, and the input is $x_1 = 1$, $x_2 =3$,
then integrated gradients attributes the change in the function value entirely to the critical variable $x_1$, whereas Shapley-Shubik assigns attributions of $1/2$ each; it seems somewhat subjective to prefer one result over the other.

We also envision other issues with applying Shapley-Shubik to deep networks:
It is computationally expensive; in an object recognition
network that takes an $100X100$ image as input, $n$ is $10000$, and
$n!$ is a gigantic number.  Even if one samples few paths randomly,
evaluating the attributions for a single path takes $n$ calls to the
deep network. In contrast, integrated gradients is able to operate
with $20$ to $300$ calls. Further, the Shapley-Shubik computation
visit inputs that are combinations of the input and the baseline. It
is possible that some of these combinations are very different from
anything seen during training. We speculate that this could lead to
attribution artifacts.
\end{remark}

\section{Applying Integrated Gradients}
\label{sec:applying}
\stitle{Selecting a Benchmark} A key step in applying integrated
gradients is to select a good baseline. \emph{We recommend that
  developers check that the baseline has a near-zero score}---as
discussed in Section~\ref{sec:method}, this allows us to interpret the
attributions as a function of the input.  But there is more to a good
baseline: For instance, for an object recogntion network it is
possible to create an adversarial example that has a zero score for a
given input label (say elephant), by applying a tiny,
carefully-designed perturbation to an image with a very different
label (say microscope) (cf. ~\cite{adversarial}). The attributions can
then include undesirable artifacts of this adversarially constructed
baseline.  So we would additionally like the baseline to convey a
complete absence of signal, so that the features that are apparent
from the attributions are properties only of the input, and not of the
baseline. For instance, in an object recognition network, a black
image signifies the absence of objects. The black image isn't unique
in this sense---an image consisting of noise has the same
property. However, using black as a baseline may result in cleaner
visualizations of ``edge'' features. For text based networks, we have
found that the all-zero input embedding vector is a good baseline. The
action of training causes unimportant words tend to have small norms,
and so, in the limit, unimportance corresponds to the all-zero
baseline. Notice that the black image corresponds to a valid input to
an object recognition network, and is also intuitively what we humans
would consider absence of signal. In contrast, the all-zero input
vector for a text network does not correspond to a valid input; it
nevertheless works for the mathematical reason described above.

\stitle{Computing Integrated Gradients}
The integral of integrated gradients can be efficiently approximated via a summation. We simply sum the gradients at points occurring at
sufficiently small intervals along the straightline path from
the baseline $\xbase$ to the input $x$.
\begin{equation}
  \begin{split}
    \integratedgrads_i^{approx}(x) \synteq \\
    (x_i-\xbase_i)\times \Sigma_{k=1}^m  \tfrac{\partial F(\xbase + \tfrac{k}{m}\times(x-\xbase)))}{\partial x_i}\times\tfrac{1}{m}
  \end{split}
\end{equation}
Here $m$ is the number of steps in the Riemman approximation of the
integral.  Notice that the approximation simply involves computing the
gradient in a for loop which should be straightforward and efficient
in most deep learning frameworks. For instance, in TensorFlow, it amounts to calling {\tt
  tf.gradients} in a loop over the set of inputs (i.e.,
$\xbase + \tfrac{k}{m}\times(x-\xbase)~\mbox{for}~ k = 1, \ldots, m$), which
could also be batched.
In practice, we find that somewhere between $20$ and $300$ steps are enough to approximate
the integral (within $5\%$); we recommend that developers \emph{check} that the attributions
approximately adds up to the difference beween the score at the input and that at the
baseline (cf. Proposition~\ref{prop:additivity}), and if not increase the step-size $m$.

\section{Applications}\label{sec:app}
The integrated gradients technique
is applicable to a variety of deep networks. Here, we
apply it to two image models, two natural language
models, and a chemistry model.

\subsection{An Object Recognition Network}\label{sec:app-incp}
We study feature attribution in an object recognition network built
using the GoogleNet architecture~\cite{SLJSRAEVR14} and trained over
the ImageNet object recognition dataset~\cite{ILSVRC15}.
We use the integrated gradients method to study pixel importance in
predictions made by this network. The gradients are computed for the
output of the highest-scoring class with respect to pixel of the input
image. The baseline input is the black image, i.e., all pixel intensities
are zero.

\begin{figure}[!htb]
  \centering
  \includegraphics[width=0.7\columnwidth]{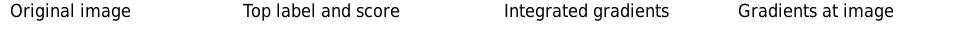}
  \includegraphics[width=0.7\columnwidth]{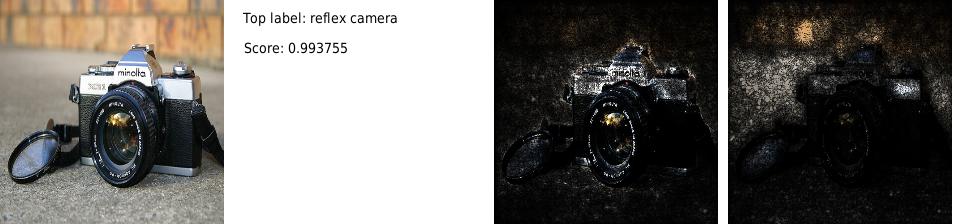}
  \includegraphics[width=0.7\columnwidth]{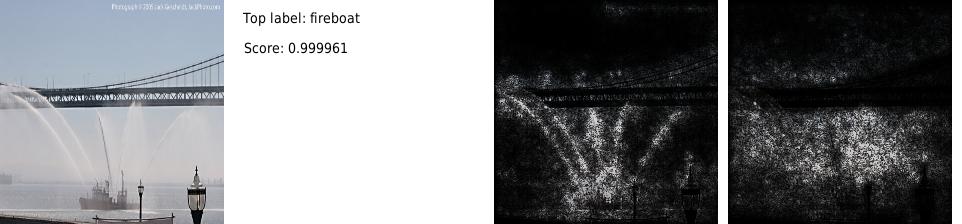}
  \includegraphics[width=0.7\columnwidth]{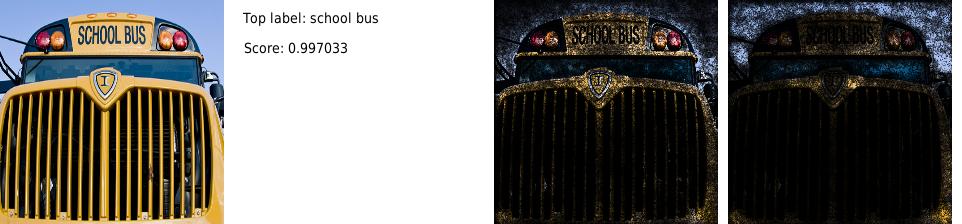}
  \includegraphics[width=0.7\columnwidth]{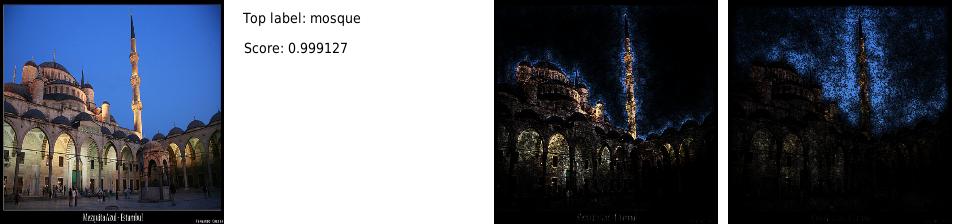}
  \includegraphics[width=0.7\columnwidth]{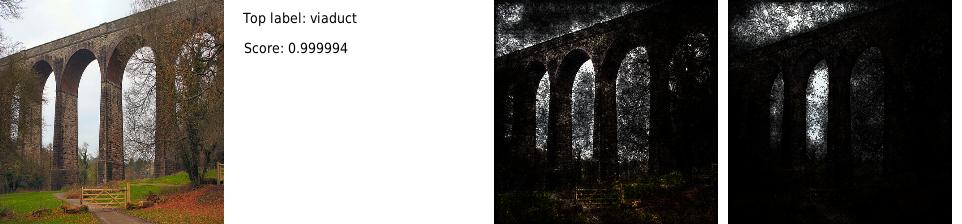}
  \includegraphics[width=0.7\columnwidth]{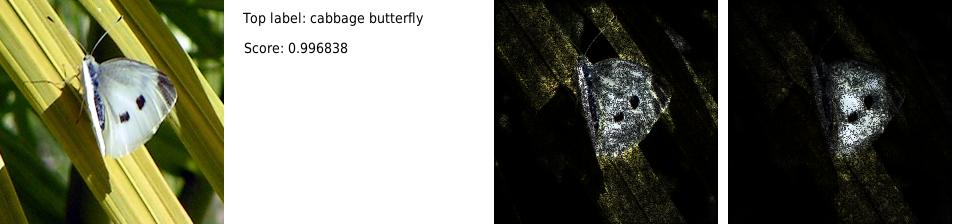}
  \includegraphics[width=0.7\columnwidth]{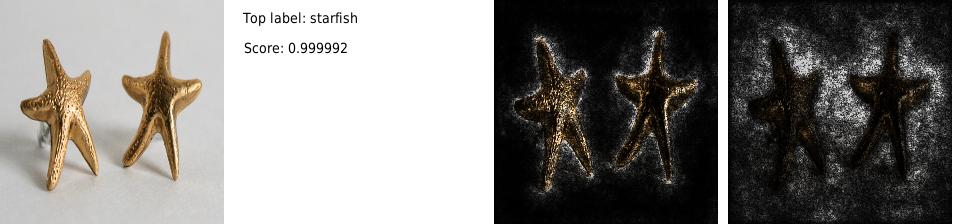}
  \caption{\textbf{Comparing integrated gradients with gradients at the image.}
    Left-to-right: original input image, label and softmax score for
    the highest scoring class, visualization of integrated gradients,
    visualization of gradients*image.
    Notice that the visualizations
    obtained from integrated gradients are better at reflecting distinctive
    features of the image.
  }\label{fig:intgrad-finalgrad}
\end{figure}

Integrated gradients can be visualized by aggregating them along the
color channel and scaling the pixels in the actual image by them.
Figure~\ref{fig:intgrad-finalgrad} shows visualizations for a
bunch of images\footnote{More examples can be found at
  \url{https://github.com/ankurtaly/Attributions}}.  For comparison,
it also presents the corresponding visualization obtained from the
product of the image with the gradients at the actual image.
Notice that integrated gradients are better at reflecting distinctive
features of the input image.


\subsection{Diabetic Retinopathy Prediction}\label{sec:app-dr}

Diabetic retinopathy (DR) is a complication of the diabetes that
affects the eyes. Recently, a deep network~\cite{jama-dr} has been proposed
to predict the severity grade for DR in retinal fundus images. The model has
good predictive accuracy on various validation datasets.

We use integrated gradients to study feature importance for this
network; like in the object recognition case, the baseline is the
black image.
Feature importance explanations are important for this network
as retina specialists may use it to
build trust in the network's predictions, decide the grade for
borderline cases, and obtain insights for further testing and
screening.

\begin{figure}[!htb]
  \centering
  \includegraphics[width=0.8\columnwidth]{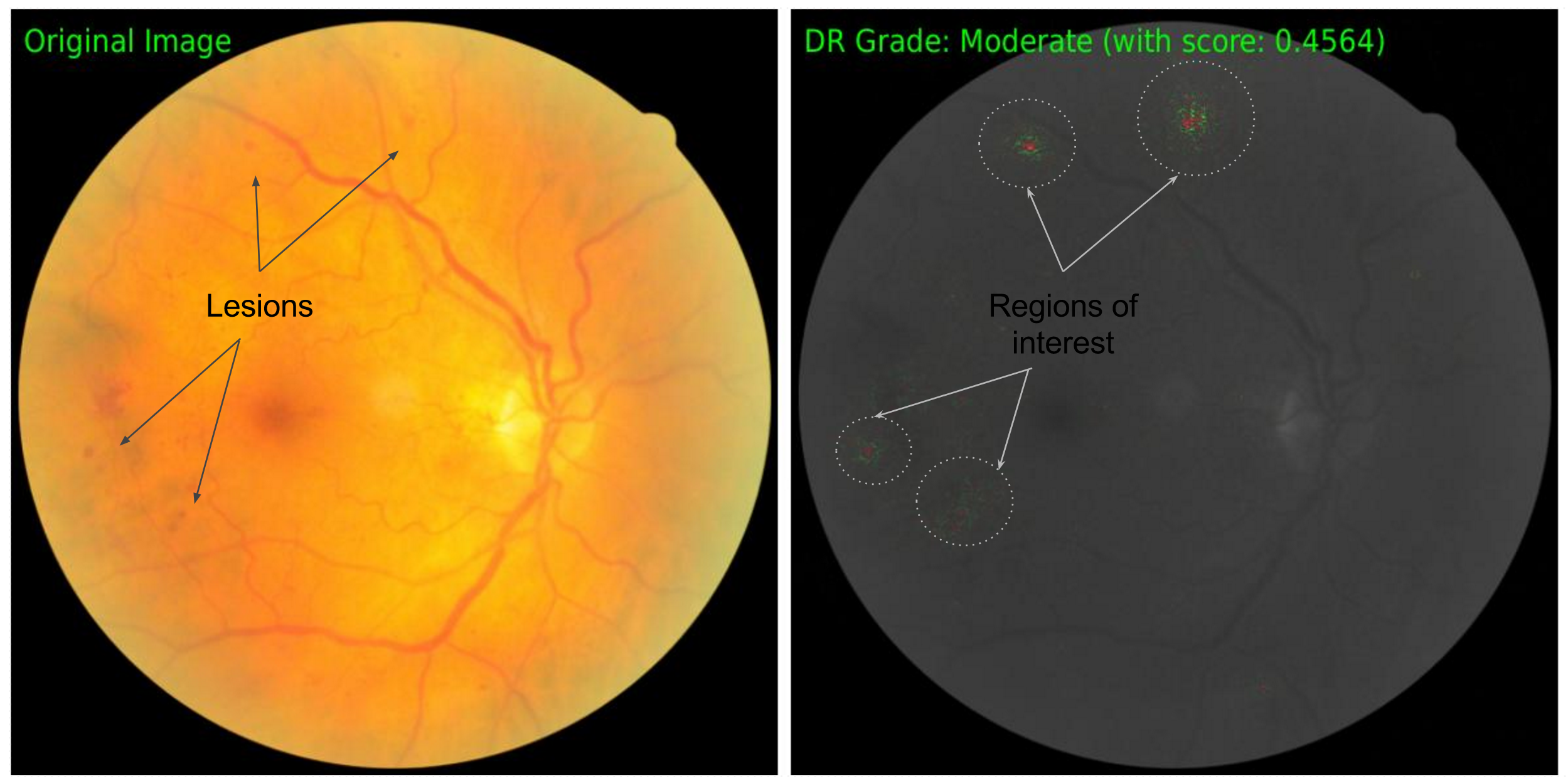}
  \caption{\textbf{Attribution for Diabetic Retinopathy grade prediction from a retinal fundus image.}
    The original image is show on the left, and the attributions (overlayed on the
    original image in gray scaee) is shown on the right. On the original image we annotate
    lesions visible to a human, and confirm that the attributions indeed point to them.}\label{fig:dr}
\end{figure}

Figure~\ref{fig:dr} shows a visualization of integrated gradients for
a retinal fundus image. The visualization method is a bit different
from that used in Figure~\ref{fig:intgrad-finalgrad}.
We aggregate integrated gradients along the color channel and
overlay them on the actual image in gray scale with positive
attribtutions along the green channel and negative attributions
along the red channel.
Notice that integrated gradients are localized
to a few pixels that seem to be lesions in the retina. The interior of
the lesions receive a negative attribution while the periphery receives
a positive attribution indicating that the network focusses on the
boundary of the lesion.


\subsection{Question Classification}\label{sec:app-qc}
Automatically answering natural language questions (over
semi-structured data) is an important problem in artificial
intelligence (AI).  A common approach is to semantically parse the
question to its logical form~\cite{L16} using a set of
human-authored grammar rules. An alternative approach is to machine
learn an end-to-end model provided there is enough training data.
An interesting question is whether one could peek inside
machine learnt models to derive new rules.
We explore this direction for a sub-problem of
semantic parsing, called \textbf{question classification}, using the
method of integrated gradients.

The goal of question classification is to identify the
type of answer it is seeking. For instance, is
the quesiton seeking a yes/no answer, or is it seeking a date? Rules
for solving this problem look for trigger phrases in the question, for
e.g., a ``when'' in the beginning indicates a date seeking
question.
We train a model for question classification using the
the text categorization architecture
proposed by~\cite{K14} over the WikiTableQuestions dataset~\cite{PL15}.
We use integrated gradients to attribute
predictions down to the question terms in order to identify new
trigger phrases for answer type.
The baseline input is the all zero embedding vector.

\begin{figure}[!htb]
  \centering
  \includegraphics[width=\columnwidth]{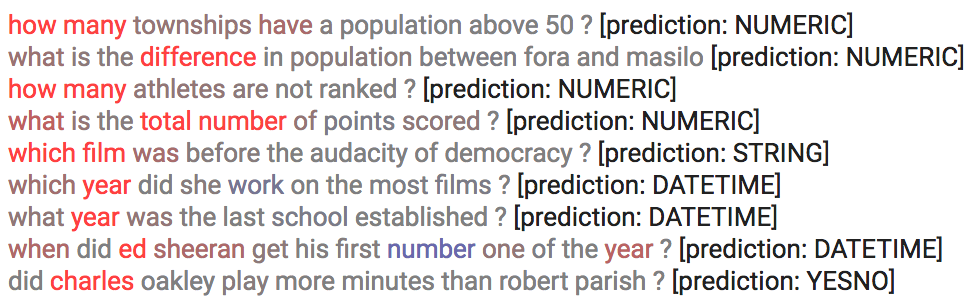}
  \caption{\textbf{Attributions from question classification model.} Term color indicates attribution
    strength---Red is positive, Blue is negative, and Gray is neutral (zero). The predicted
    class is specified in square brackets.}
  \label{fig:qc}
\end{figure}

Figure~\ref{fig:qc} lists a few questions with constituent terms
highlighted based on their attribution.  Notice that the attributions
largely agree with commonly used rules, for e.g., ``how many''
indicates a numeric seeking question. In addition, attributions
help identify novel question classification rules, for e.g.,
questions containing ``total number'' are seeking numeric answers.
Attributions also point out undesirable correlations, for e.g.,
``charles'' is used as trigger for a yes/no question.

\subsection{Neural Machine Translation}\label{sec:app-nmt}
\begin{figure}[!htb]
  \centering
  \includegraphics[width=0.7\columnwidth]{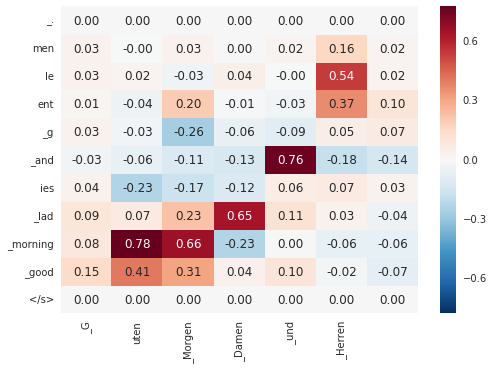}
  \caption{\textbf{Attributions from a language translation model.} Input in English: ``good morning ladies and gentlemen''. Output in German: ``Guten Morgen Damen und Herren''. Both input and output are tokenized into word pieces, where a word piece prefixed by underscore indicates that it should be the prefix of a word.}
  \label{fig:nmt}
\end{figure}

We applied our technique to a complex, LSTM-based Neural Machine Translation System~\cite{Wu16}.
We attribute the output probability of every output token (in form of wordpieces)
to the input tokens. Such attributions ``align'' the output sentence with the input sentence.
For baseline, we zero out the embeddings of all tokens except the start and end markers.
Figure~\ref{fig:nmt} shows an example of such an attribution-based alignments.
We observed that the results make intuitive sense. E.g. ``und'' is mostly attributed to ``and'',
and ``morgen'' is mostly attributed to ``morning''.
We use $100-1000$ steps (cf. Section~\ref{sec:applying}) in the integrated gradient approximation; we need this because the network is highly nonlinear.

\subsection{Chemistry Models}\label{sec:app-chem}
We apply integrated gradients to a network performing Ligand-Based
Virtual Screening which is the problem of predicting whether an input
molecule is active against a certain target (e.g., protein or enzyme).
In particular, we consider a network based on the molecular graph
convolution architecture proposed by~\cite{KMBPR16}.

The network requires an input
molecule to be encoded by hand as a set of atom and atom-pair features
describing the molecule as an undirected graph. Atoms are featurized
using a one-hot encoding specifying the atom type (e.g., C, O, S,
etc.), and atom-pairs are featurized by specifying either the type of
bond (e.g., single, double, triple, etc.)  between the atoms, or the
graph distance between them.
The baseline input is obtained zeroing out the feature vectors for
atom and atom-pairs.

\begin{figure}[!ht]
  \centering
  \includegraphics[width=0.9\columnwidth]{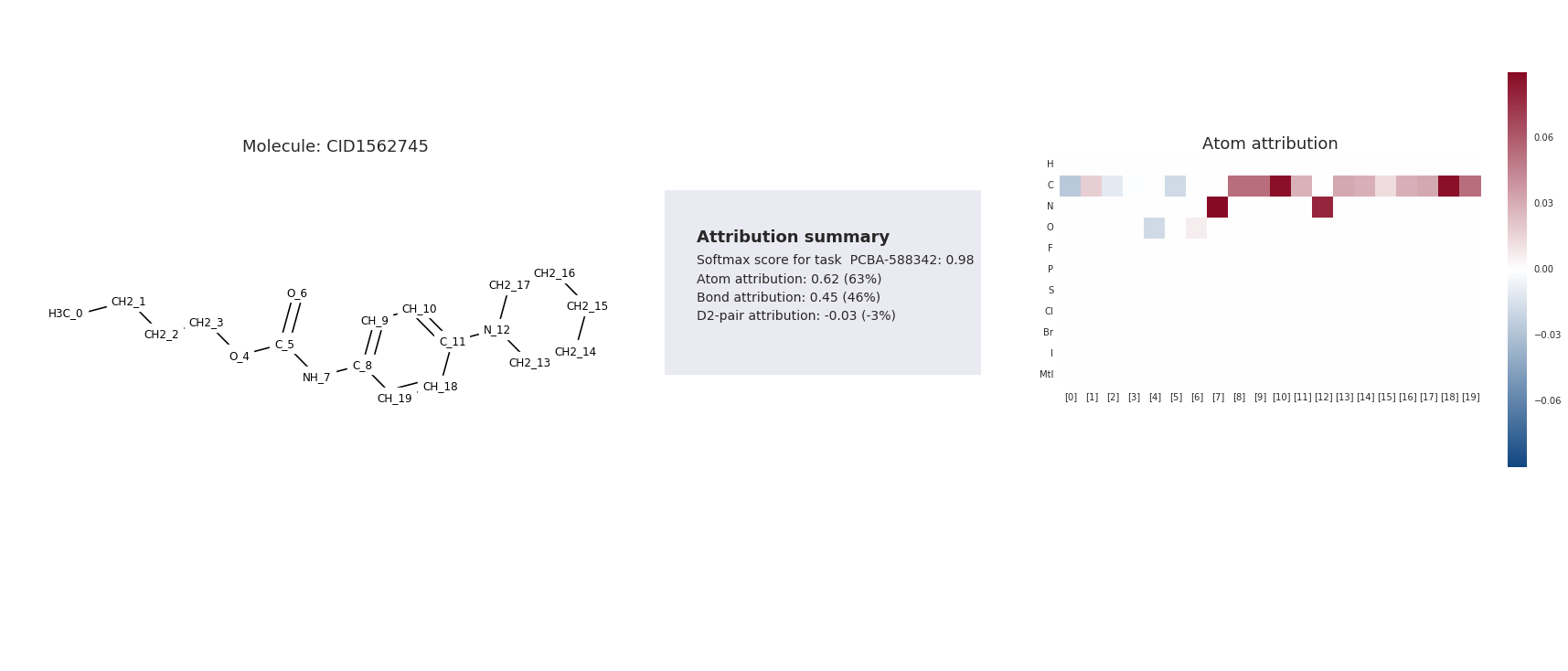}
  \includegraphics[width=0.9\columnwidth]{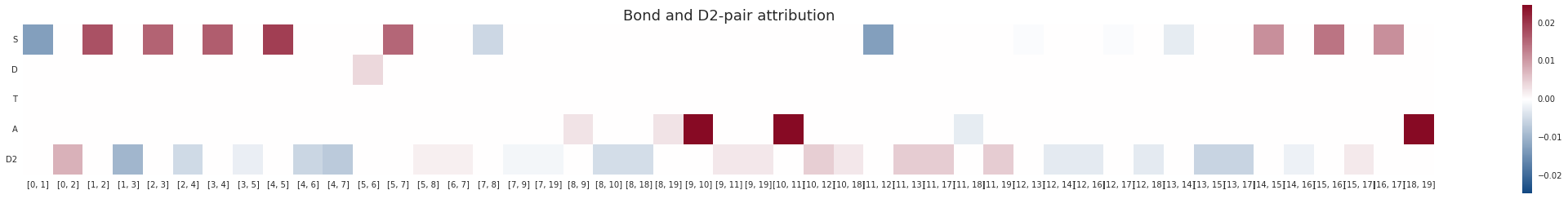}
  \caption{\textbf{Attribution for a molecule under the W2N2 network~\cite{KMBPR16}}.
    The molecules is active on task PCBA-58432.}\label{fig:attribution-gas}
\end{figure}

We visualize integrated gradients as heatmaps over the the atom and
atom-pair features with the heatmap intensity depicting the strength
of the contribution. Figure~\ref{fig:attribution-gas} shows the
visualization for a specific molecule. Since integrated gradients add
up to the final prediction score (see
Proposition~\ref{prop:additivity}), the magnitudes can be use for
accounting the contributions of each feature.  For instance, for the
molecule in the figure, atom-pairs that have a bond between them
cumulatively contribute to $46\%$ of the prediction score, while all
other pairs cumulatively contribute to only $-3\%$.


\stitle{Identifying Degenerate Features}
We now discuss how attributions helped us spot an anomaly in the W1N2
architecture in ~\cite{KMBPR16}. On applying the integrated gradients
method to this network, we found that several atoms in the same
molecule received identical attribution despite being bonded to
different atoms. This is surprising as one would expect two atoms with
different neighborhoods to be treated differently by the network.

On investigating the problem further, in the network architecture,
the atoms and atom-pair features were not fully convolved. This
caused all atoms that have the
same atom type, and same number of bonds of each type to contribute
identically to the network.

\section{Other Related work}\label{sec:related}
We already covered closely related work on attribution in
Section~\ref{sec:two-axioms}.  We mention other related work.
Over the last few years, there has been a vast amount work on
demystifying the inner workings of deep networks. Most of this work
has been on networks trained on computer vision tasks, and deals with
understanding what a specific neuron computes~\cite{EBCV09, QVL13}
and interpreting the representations captured by neurons during a
prediction~\cite{MV15, DB15, YCNFL15}. In contrast, we focus on
understanding the network's behavior on a specific input in terms of
the base level input features. Our technique quantifies the importance
of each feature in the prediction.

 One approach to the attribution problem proposed first
 by~\cite{RSG16a, RSG16b}, is to locally approximate the behavior of
 the network in the vicinity of the input being explained with a
 simpler, more interpretable model.  An appealing aspect of this
 approach is that it is completely agnostic to the implementation of
 the network and satisfies implemenation invariance. However, this
 approach does not guarantee sensitivity. There is no guarantee that
 the local region explored escapes the ``flat'' section of the
 prediction function in the sense of Section~\ref{sec:two-axioms}. The other
 issue is that the method is expensive to implement for networks with
 ``dense'' input like image networks as one needs to explore a local
 region of size proportional to the number of pixels and train a model
 for this space. In contrast, our technique works with a few calls to
 the gradient operation.

Attention mechanisms~\cite{BahdanauCB14} have gained popularity recently. One may think that
attention could be used a proxy for attributions, but this has issues. For instance, in a LSTM that also employs attention, there are many ways for an input token to influence an output token: the memory cell, the recurrent state, and ``attention''.
Focussing only an attention ignores the other modes of influence and results in an incomplete picture.



\section{Conclusion}

The primary contribution of this paper is a method called integrated
gradients that attributes the prediction of a deep network to its
inputs. It can be implemented using a few calls to the gradients
operator, can be applied to a variety of deep networks, and has a
strong theoretical justification.

A secondary contribution of this paper is to clarify desirable
features of an attribution method using an axiomatic framework
inspired by cost-sharing literature from economics. Without the
axiomatic approach it is hard to tell whether the 
attribution method is affected by data artifacts, network's artifacts or
artifacts of the method. The axiomatic approach rules out
artifacts of the last type.

While our and other works have made some progress on understanding
the relative importance of input features in a deep network,
we have not addressed the interactions between the input features or
the logic employed by the network. So there remain many unanswered
questions in terms of debugging the I/O behavior of a deep
network.



\subsubsection*{Acknowledgments}
We would like to thank Samy Bengio, Kedar Dhamdhere, Scott Lundberg, Amir Najmi, Kevin McCurley, Patrick Riley, Christian Szegedy, Diane Tang for their
feedback. We would like to thank Daniel Smilkov and Federico Allocati for identifying bugs in our descriptions.
We would like to thank our anonymous reviewers for identifying bugs, and their suggestions to improve presentation.

\bibliography{paper-icml}
\bibliographystyle{icml2017}
\appendix

\section{Proof of Theorem~\ref{thm:unique}}\label{sec:proof}
\begin{proof}
  Consider a non-straightline path $\pathfn:[0,1]\to \reals^n$ from baseline
  to input.  W.l.o.g.,
there exists $t_0\in [0,1]$ such that for two dimensions $i,j$,
$\pathfn_i(t_0) > \pathfn_j(t_0)$.  Let $(t_1, t_2)$ be the maximum real
open interval containing $t_0$ such that $\pathfn_i(t) > \pathfn_j(t)$
for all $t$ in $(t_1, t_2)$, and let $a=\pathfn_i(t_1)=\pathfn_j(t_1)$,
and $b=\pathfn_i(t_2)=\pathfn_j(t_2)$.  Define function $f:x\in
[0,1]^n\to R$ as $0$ if $\min(x_i, x_j) \leq a$, as $(b - a)^2$ if
$\max(x_i,x_j)\geq b$, and as $(x_i - a)(x_j-a)$ otherwise.  Next we
compute the attributions of $f$ at $x=\langle 1,\ldots,1\rangle_n$ with
baseline $\xbase=\langle 0,\ldots,0\rangle_n$.
Note that $x_i$ and $x_j$ are symmetric, and should get identical
attributions.  For $t\notin [t_1, t_2]$, the function is a constant,
and the attribution of $f$ is zero to all variables, while for
$t\in(t_1, t_2)$, the integrand of attribution of $f$ is $\pathfn_j(t) - a$ to
$x_i$, and $\pathfn_i(t) - a$ to $x_j$, where the latter is always
strictly larger by our choice of the interval. Integrating, 
it follows that $x_j$ gets a larger attribution than $x_i$, contradiction.
\end{proof}

\section{Attribution Counter-Examples}\label{sec:examples}
We show that the methods DeepLift and Layer-wise relevance propagation
(LRP) break the implementation invariance axiom, and the Deconvolution
and Guided back-propagation methods break the sensitivity axiom.
\begin{figure}[!htb]
  \centering
  \begin{subfigure}{.9\columnwidth}
    \includegraphics[width=0.9\columnwidth]{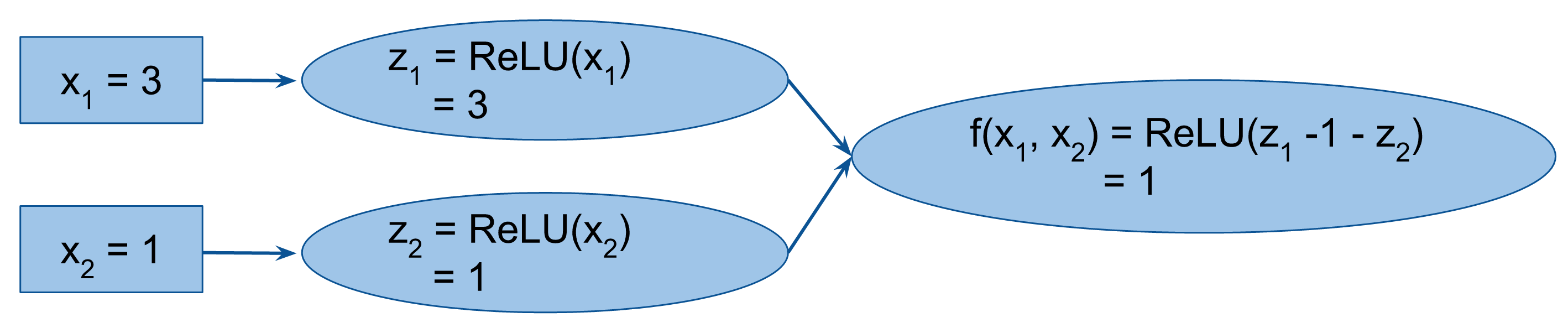}    
    \tiny
    \caption*{\footnotesize
      Network $f(x_1, x_2)$\\
      Attributions at $x_1 = 3, x_2 = 1$\\
      $\begin{array}{ll}
        \mbox{\bf Integrated gradients} & x_1 = 1.5,~x_2 = -0.5 \\
        \mbox{DeepLift} & x_1 = 1.5,~x_2 = -0.5 \\
        \mbox{LRP} & x_1 = 1.5,~x_2 = -0.5 \\
       \end{array}$
    }\label{fig:deeplift-1}
  \end{subfigure}
  \\
  \begin{subfigure}{.9\columnwidth}
    \includegraphics[width=0.9\columnwidth]{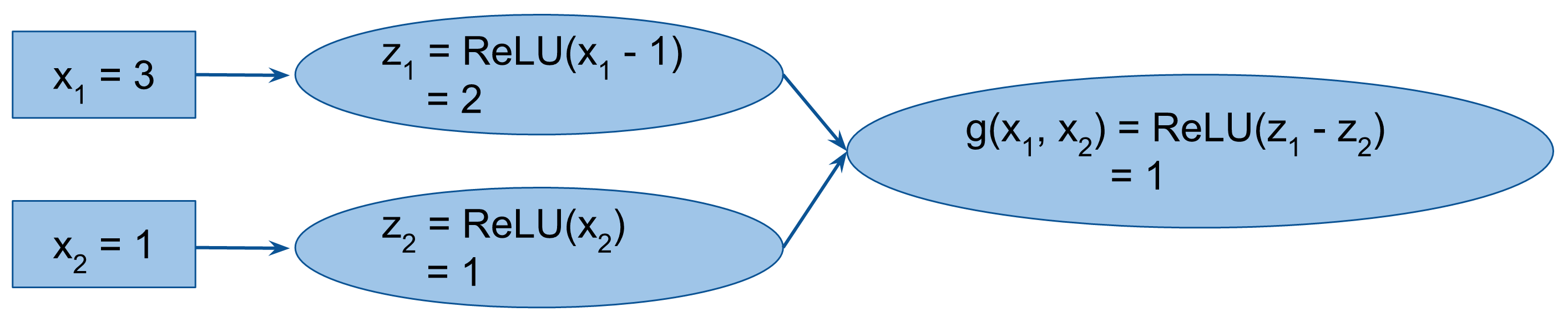}
    \caption*{\footnotesize
      Network $g(x_1, x_2)$\\
      Attributions at $x_1 = 3, x_2 = 1$\\
      $\begin{array}{ll}
        \mbox{\bf Integrated gradients} & x_1 = 1.5,~x_2 = -0.5 \\
        \mbox{DeepLift} &  x_1 = 2,~x_2 = -1 \\
        \mbox{LRP} & x_1 = 2,~x_2 = -1 \\
       \end{array}$
    }\label{fig:deeplift-2}
  \end{subfigure}
  \caption{\textbf{Attributions for two functionally equivalent
      networks}. The figure shows attributions for two functionally
    equivalent networks $f(x_1, x_2)$ and $g(x_1, x_2)$ at the input
    $x_1 = 3,~x_2 = 1$ using integrated gradients, DeepLift
   ~\cite{SGSK16}, and Layer-wise relevance propagation (LRP)
   ~\cite{BMBMS16}. The reference input for Integrated gradients and
    DeepLift is $x_1 = 0,~x_2 = 0$. All methods except integrated
    gradients provide different attributions for the two
    networks.}\label{fig:deeplift}
\end{figure}

Figure~\ref{fig:deeplift} provides an example of
two equivalent networks $f(x_1, x_2)$ and $g(x_1, x_2)$ for which DeepLift and LRP
yield different attributions.

First, observe that the networks $f$ and $g$ are of the form
$f(x_1, x_2) = \relu(h(x_1, x_2))$ and $f(x_1, x_2) = \relu(k(x_1, x_2))$\footnote{
  $\relu(x)$ is defined as $\ms{max}(x, 0)$.},
where
$$\begin{array}{l}
  h(x_1, x_2) = \relu(x_1) - 1 - \relu(x_2) \\
  k(x_1, x_2) = \relu(x_1 - 1) - \relu(x_2)
\end{array}$$
Note that $h$ and $k$ are not equivalent. They have different values whenever $x_1<1$.
But $f$ and $g$ are equivalent. To prove this, suppose for contradiction that
$f$ and $g$ are different for some $x_1,x_2$. Then it must be the case that
$\relu(x_1) - 1 \neq \relu(x_1 - 1)$. This happens only when $x_1<1$,
which implies that $f(x_1, x_2)=g(x_1, x_2)=0$.

Now we leverage the above example to show that Deconvolution and
Guided back-propagation break sensitivity.  Consider the network
$f(x_1, x_2$) from Figure~\ref{fig:deeplift}.  For a fixed value of
$x_1$ greater than $1$, the output decreases linearly as $x_2$
increases from $0$ to $x_1 - 1$. Yet, for all inputs, Deconvolutional
networks and Guided back-propagation results in zero attribution for
$x_2$. This happens because for all inputs the back-propagated signal
received at the node $\relu(x_2)$ is negative and is therefore not
back-propagated through the $\relu$ operation (per the rules of
deconvolution and guided back-propagation; see~\cite{SDBR14} for
details). As a result, the feature $x_2$ receives zero attribution
despite the network's output being sensitive to it.

\end{document}